\documentclass[times]{ouparticle}

\usepackage{amsmath}
\usepackage{amssymb}
\usepackage{graphicx}
\usepackage{hyperref}
\usepackage{float}
\usepackage{enumitem}
\usepackage{multicol}

\begin{document}

\title{Spectrum: Targeted Training on Signal to Noise Ratio}

\author{%
\name{Eric Hartford*}
\address{Cognitive Computations}
\and
\name{Lucas Atkins*}
\address{Arcee.AI}
\and
\name{Fernando Fernandes Neto*}
\address{Cognitive Computations}
\and
\name{David Golchinfar}
\address{Vago Solutions}}

\abstract{
Efficiently post-training large language models remains a challenging task due to the vast computational resources required. We present Spectrum, a method that accelerates LLM training by selectively targeting layer modules based on their signal-to-noise ratio (SNR), and freezing the remaining modules. Our approach, which utilizes an algorithm to compute module SNRs prior to training, has shown to effectively match the performance of full fine-tuning while reducing GPU memory usage. Experiments comparing Spectrum to existing methods such as QLoRA demonstrate its effectiveness in terms of model quality and VRAM efficiency in distributed environments.\\}


\maketitle

\section{Introduction}

Large language models (LLMs) have showcased impressive abilities across various natural language tasks. Efficiently training these massive models remains a major challenge, demanding extensive computational resources and time. To address this issue, a growing body of research has focused on developing techniques to minimize the memory footprint and accelerate LLM training without sacrificing performance.
\newline

We present Spectrum, a method for selectively training the layers of an LLM based on their signal-to-noise ratio (SNR). Rooted in Random Matrix Theory, Spectrum utilizes the Marchenko-Pastur distribution to pinpoint informative layers according to their SNR. Unlike previous approaches, such as QLoRA \cite{dettmers2023qlora}, which quantizes the entire model, Spectrum strategically targets specific layers and modules for training while keeping others frozen. By concentrating computational resources on the most informative parameters, Spectrum achieves superior performance while significantly reducing training time and memory requirements compared to state-of-the-art methods.
\newline

Our main contributions are as follows:

\begin{itemize}

\item We propose Spectrum, an approach for efficient LLM training that selectively trains layers based on their SNR. 

\item We conduct experiments comparing Spectrum to Full Finetuning and QLoRA on a range of language benchmarks. Spectrum consistently matches or outperforms prior methods training faster while using less GPU memory.

\item We provide analysis of why Spectrum is so effective, demonstrating that focusing on high-SNR layers enables more efficient use of training compute.

\item We release our code publicly to facilitate future research.

\end{itemize}

Spectrum opens up exciting opportunities to train large language models cost-effectively without compromising quality. This has significant implications for democratizing LLM research and enabling new applications.

\section{Related Work}

Efficient training of large language models has attracted significant attention. Two notable prior works that aim to reduce the computational cost and memory requirements of LLM training and deployment are QLoRA \cite{dettmers2023qlora} and LASER \cite{sharma2023truth}. Dettmers et al. \cite{dettmers2023qlora} introduced QLoRA, a method that combines LoRA (Low-Rank Adaptation) \cite{hu2021lora} with 4-bit quantization to enable efficient finetuning of LLMs. By quantizing the base model weights to 4 bits and storing the LoRA adaptation parameters in 16-bit precision, QLoRA significantly reduces memory usage during training. This allows finetuning of billion-parameter models on a single GPU. However, QLoRA applies quantization and LoRA uniformly across all model layers, which may not be optimal. LASER \cite{sharma2023truth} selectively applies a low-rank approximation to specific layers of a trained model. The authors measure the signal-to-noise ratio (SNR) of each layer and reduce the rank of high-SNR layers, showing that this can boost performance on certain downstream tasks. LASER is primarily designed for model compression after training rather than improving training itself.
\newline

Spectrum builds upon the insights of QLoRA and LASER while addressing their limitations. Like QLoRA, Spectrum enables efficient training of language models. 
Instead of quantizing all layers, Spectrum selectively trains a subset of layers in full precision based on their SNR. This allows devoting compute to the most informative parameters during training.  Similar to LASER, Spectrum uses SNR to identify important layers. But while LASER approximates layers after training for model \textit{compression}, Spectrum selects layer modules to \textit{train} based on their SNR. Spectrum combines the strengths of QLoRA and LASER - efficient memory usage and SNR-based layer selection - while improving upon them to enable faster, better LLM training.

\section{Mathematical Foundation}
Our approach is grounded in Random Matrix Theory (RMT), enabling efficient identification of the most informative layers for targeted training. We utilize the Marchenko-Pastur distribution \cite{marchenko1967distribution}, which characterizes eigenvalue distributions in large random matrices, to distinguish signal from noise in the network's weight matrices. In this section some key aspects of mathematical foundation of the method are provided, such as SVD, Marchenko-Pastur Distributions and  learning representations with infrequent/noisy terms often lead to singular values close to zero.
\newline

Begining with Singular Value Decompositions (SVD), it is important to elucidate that it enables to assess part of knowledge representation in matrices. Lower singular values often signify noise, less important information, or less frequent terms in data. Zeroing these values, as demonstrated in the LASER paper \cite{sharma2023truth}, acts as a denoising process, enhancing the quality of learned representations. This process, however, must be careful not to suppress factual knowledge or impact the learning of more complex patterns, which can lead to catastrophic forgetting.

\subsection{Illustrating Overfitting's Impact on Singular Values}
Consider a dataset with $n$ data points $(x_i, y_i)$. In polynomial regression of degree $d$, the design matrix $X$ is constructed as:
\begin{align*}
X = \begin{bmatrix}
1 & x_1 & x_1^2 & \cdots & x_1^d \\
1 & x_2 & x_2^2 & \cdots & x_2^d \\
\vdots & \vdots & \vdots & & \vdots \\
1 & x_n & x_n^2 & \cdots & x_n^d
\end{bmatrix}
\end{align*}

Overfitting occurs when the model is too complex relative to the data, specifically in polynomial regression when the degree $d$ is high compared to the number of data points $n$. This results in linear dependency among columns of $X$, leading to a rank-deficient matrix with zero singular values.

The singular value decomposition (SVD) of $X$ is given by:
\begin{align*}
X = U \Sigma V^T
\end{align*}
where $U, V$ are orthogonal matrices, and $\Sigma$ is a diagonal matrix of singular values. Linear dependency in $X$ results in zero singular values in $\Sigma$, impacting the stability and reliability of regression coefficients calculated.

A design matrix $X$ with zero singular values leads to issues in solving the normal equation for polynomial regression, $X^T X \beta = X^T y$, due to the non-invertibility of $X^T X$. This directly contributes to overfitting, indicating that the model has fitted the noise instead of the underlying data pattern.

\subsection{Random Matrix Theory (RMT) Perspective}
RMT provides insights into the nature of data represented by a matrix. For large matrices from real-world data, the bulk of eigenvalues/singular values typically forms a ``bulk spectrum". Values associated with less frequent data points often deviate from this bulk and can be misinterpreted as noise. RMT helps distinguish signal from noise, but identifying meaningful signals from deviations requires careful consideration.

\subsection{Benefits of Focusing on Matrices with Larger Singular Values}
Skipping matrices with insignificant singular values has several advantages: preservation of factual and scattered information from the pre-training phase, retaining layers with diverse information; training focuses on more stable and well-posed matrices with larger max-min singular values; targeting matrices with larger singular values enables us to emphasize transformations with the largest impact on latent representations.

\subsection{Relating Eigenvalues and Singular Values}
For a weight matrix $W$ in a neural network, the SVD is given by:
\begin{align*}
W = USV^T
\end{align*}
where $U, V$ are orthogonal, and $S$ is a diagonal matrix of singular values. The eigenvalues of $W^TW$ are related to the squared singular values of $W$:
\begin{align*}
W^TW = VS^2V^T
\end{align*}

\subsection{Marchenko-Pastur Distribution}
The Marchenko-Pastur distribution \cite{marchenko1967distribution} describes the eigenvalue distribution of large random matrices as dimensions tend to infinity with a fixed aspect ratio. The main insight from the previous subsection regarding applying the Marchenko-Pastur distribution is that this is applicable only to square matrices. Hence, when squaring a rectangular matrix, allows one to obtain a square matrix, and hence, applying the Marchenko-Pastur distributions for its eigenvalues. Thus, the relationship between eigenvalues of $W^TW$ and squared singular values of $W$ is central to applying the Marchenko-Pastur distribution. 
For a matrix $W$ of size $m \times n$, the eigenvalues of $C = (1/n)W^TW$ converge to a distribution bounded by:
\begin{align*}
\lambda_+ = \sigma^2\left(1 + \sqrt{\frac{m}{n}}\right)^2, \\
\lambda_- = \sigma^2\left(1 - \sqrt{\frac{m}{n}}\right)^2
\end{align*}
where $\lambda_+, \lambda_-$ are the largest and smallest eigenvalues, and $\sigma$ is the standard deviation. This leads to bounds on the singular values of $W$:
\begin{align*}
\varepsilon_+ = \frac{1}{\sqrt{n}}\sigma\left(1 + \sqrt{\frac{m}{n}}\right), \\
\varepsilon_- = \frac{1}{\sqrt{n}}\sigma\left(1 - \sqrt{\frac{m}{n}}\right)
\end{align*}

\subsection{Signal-to-Noise Ratio and Matrix Ranking}
To ensure numerical stability and efficient computation, we omit the normalization term $(1/\sqrt{n})$ when calculating singular value bounds. We also use the interquartile range instead of the standard deviation to account for potential skewness and kurtosis. The signal-to-noise ratio (SNR) of a weight matrix is defined as:
\begin{align*}
SNR = \frac{\sum_{k | \sigma_k > \varepsilon} \sigma_k}{\sum_{n | \sigma_n < \varepsilon} \sigma_n}
\end{align*}
where $\varepsilon$ separates signal from noise singular values. We normalize $SNR$ by the largest singular value for sensitivity analysis, enhanced comparison, and conditioning information. Matrices with higher SNR contain more informative features and less noise, making them ideal targets for efficient learning and improved model performance.

\section{Measuring The Signal-to-Noise Ratio}
Our Spectrum tool computes layer SNRs. For each layer, Spectrum computes the SVD of the weight matrix, calculates the SNR, and normalizes it by the largest singular value. The noise threshold $\varepsilon$ is determined using the Marchenko-Pastur distribution, effectively separating signal from noise. The SNR formula aligns with the theoretical understanding of the singular value spectrum partitioned by the Marchenko-Pastur distribution.
Spectrum computes SNRs in batches, efficiently analyzing the entire model. By leveraging RMT and the Marchenko-Pastur distribution, Spectrum provides a principled approach to measuring layer SNRs, forming the basis for selective training.

\subsection{Layer Selection}
Spectrum selects layers with higher SNRs, containing more task-relevant information, for targeted updates. The number of layers trained is a hyperparameter balancing training speed and performance. By default, Spectrum trains the top 25\% of layers in each module, ensuring a balanced distribution of updates.
\section{Evaluations}

We present an evaluation of Spectrum on a set of language model evaluations. We compare Spectrum to full finetuning and QLoRA \cite{dettmers2023qlora} in terms of training speed, memory usage, and benchmark results. We use Spectrum-50 and Spectrum-25 for these evaluations

\subsection{Setup}

We trained five Llama 3 8B\footnote{\url{https://huggingface.co/meta-llama/Meta-Llama-3-8B}} models using airoboros-3.1\footnote{\url{https://huggingface.co/datasets/jondurbin/airoboros-3.1}} as the dataset. The models were trained as such: one with a full finetune, another with QLoRA, one targeting the top 50\% SNR layers with Spectrum, and another targeting the top 25\%  We include a fifth model trained with QLoRA that targets the top 25\% layer modules identified by Spectrum. We chose airoboros to establish a baseline due to its relatively small size but large number of general language understanding tasks.
\newline

We run the same training procedure (excluding QLoRA + Spectrum) on Mistral 7B\footnote{\url{https://huggingface.co/mistralai/Mistral-7B-v0.1}}. Each model was initialized from the base models released by Meta and Mistral. All models were trained for two epochs with the same hyperparameters: a learning rate of 1e-5, gradient norm of 4, batch size of 1, and a maximum sequence length of 4096. The only change made was during QLoRA training where a lower learning rate of 2e-4 was used.
\newline

All experiments were conducted on an 8xL40S (46GB VRAM per GPU) node provided by Crusoe Energy. We used Hugging Face Accelerate and Axolotl with DeepSpeed ZeRO-3 \cite{aminabadi2022deepspeed} for distributed training. To compare performance on single GPU jobs, we retrained our llama-3-8b models on a 1x Nvidia L40S GPU for a single epoch to assess single GPU VRAM usage. Our QLoRA hyperparameters within Axolotl were as follows:

\begin{center}
\begin{minipage}{0.5\linewidth}
\begin{verbatim}
adapter: qlora
lora_r: 32
lora_alpha: 16
lora_dropout: 0.05
lora_target_linear: true
lora_modules_to_save: [embed_tokens, lm_head]
\end{verbatim}
\end{minipage}
\end{center}

\subsection{Benchmark Scores}

\begin{figure}[H]
\centering\
\caption{}
\includegraphics[width=1\linewidth]{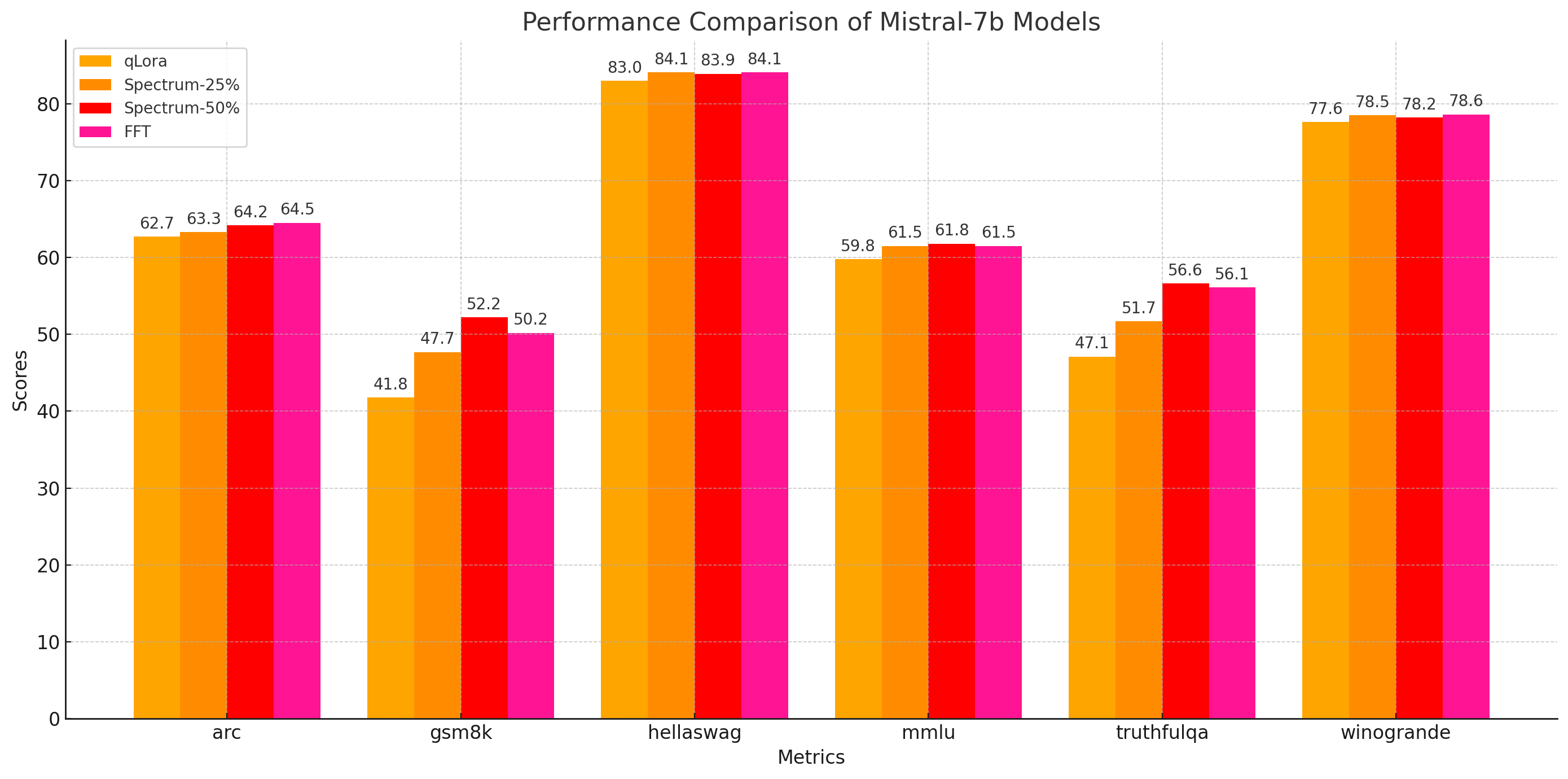}
\label{fig:mistral-7b-ollm}
\end{figure}

\begin{figure}[H]
\centering
\caption{}
\includegraphics[width=1\linewidth]{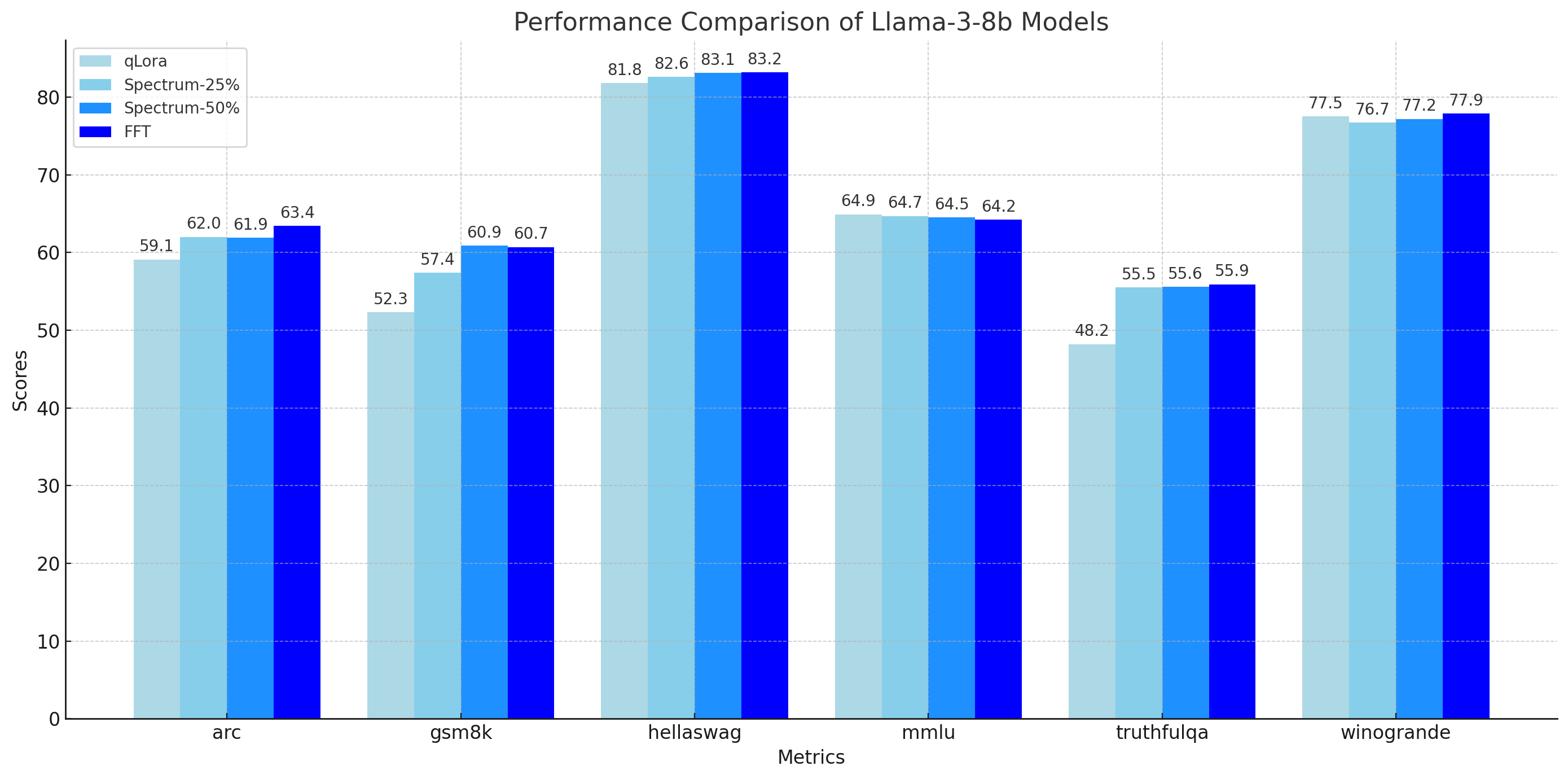}
\label{fig:llama-3-ollm}
\end{figure}

\begin{figure}[H]
\centering
\caption{}
\includegraphics[width=1\linewidth]{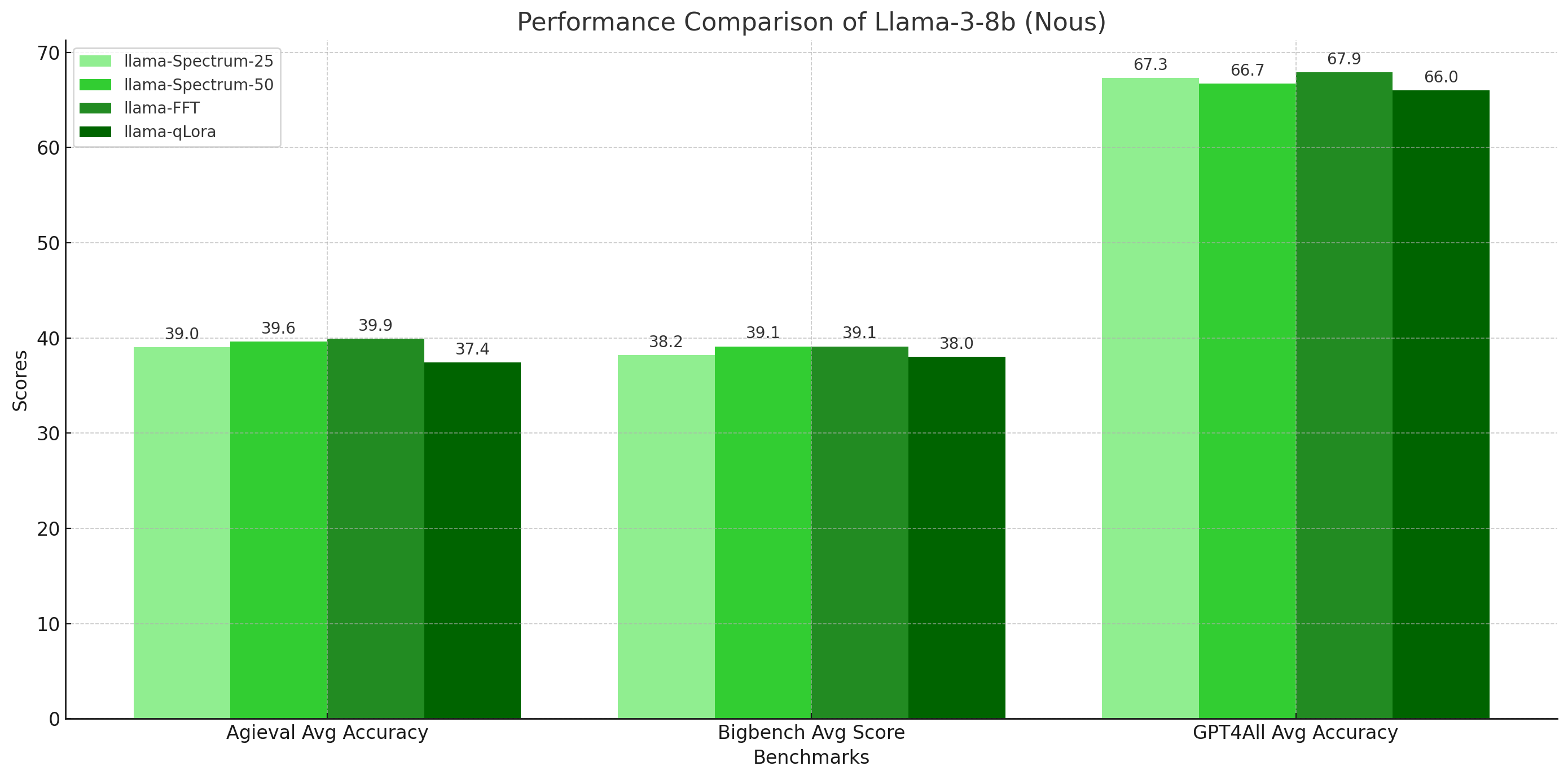}
\label{fig:enter-label}
\end{figure}

\subsubsection{Memory Usage \& Training Time}
We measured the memory usage and training time for different training configurations. Models are compared against baseline FFT training in terms of peak memory usage per GPU, VRAM usage on a single GPU, and total training time.
\begin{table}[H]
\centering
\renewcommand{\arraystretch}{1.3}
\label{table:distributed_training}
\caption{Distributed Training}
\resizebox{\textwidth}{!}{
\begin{tabular}{|l|c|c|}
\hline
Model & Peak Memory Usage per GPU & \% Efficiency Compared to FFT \\
\hline
Llama-3-8b-FFT & 24.92 GB & Baseline \\
Llama-3-8b-QLoRA & 21.25 GB & 14.73\% \\
Llama-3-8b-Spectrum-50 & 20.50 GB & 17.72\% \\
Llama-3-8b-Spectrum-25 & 19.18 GB & 23.05\% \\
Llama-3-8b-Spectrum-25+QLoRA & 16.95 GB & 31.99\% \\
\hline
\end{tabular}
}
\end{table}

\begin{table}[H]
\centering
\renewcommand{\arraystretch}{1.3}
\begin{minipage}{0.48\textwidth}
\centering
\resizebox{\textwidth}{!}{
\begin{tabular}{|l|c|}
\hline
Model & Single GPU VRAM Usage \\
\hline
Llama-3-8b-FFT & N/A (out of memory) \\
Llama-3-8b-Spectrum-50 & 34.65 GB \\
Llama-3-8b-Spectrum-25 & 27.46 GB \\
Llama-3-8b-QLoRA & 23.39 GB \\
Llama-3-8b-Spectrum-25+QLoRA & 21.18 GB \\
\hline
\end{tabular}
}
\caption{Single GPU VRAM Usage}
\label{table:single_gpu_vram}
\end{minipage}
\hfill
\begin{minipage}{0.48\textwidth}
\centering
\resizebox{\textwidth}{!}{
\begin{tabular}{|l|c|}
\hline
Model & Training Time (8xL40S) \\
\hline
Llama-3-8b-FFT & 1h 43m 16s \\
Llama-3-8b-Spectrum-50 & 1h 27m 17s \\
Llama-3-8b-QLoRA & 1h 18m 14s \\
Llama-3-8b-Spectrum-25 & 1h 5m 33s \\
Llama-3-8b-Spectrum-25+QLoRA & 54m 55s \\
\hline
\end{tabular}
}
\caption{Training Time}
\label{table:training_time}
\end{minipage}
\end{table}
\clearpage

\subsection{Analysis}
In our tests, Spectrum not only competes with fully fine-tuned models but also, in some cases, outperforms them in terms of benchmark scores. Additionally, Spectrum surpasses QLoRA across all metrics except for single GPU VRAM usage. Spectrum-50, which targets the top 50\% of layers, matches or surpasses full finetuning in various benchmarks. This indicates that focusing on high SNR layers is sufficient for competitive performance. Spectrum-25, targeting the top 25\% of layers, also shows strong results, at times scoring higher than full finetuning and Spectrum-50. This suggests that critical post-training information resides in a small subset of layers. Our experiments with Spectrum-25+QLoRA showed promising evaluation results compared to QLoRA alone, with significant reductions in VRAM use.

\begin{figure} [H]
    \centering
    \caption{}
    \includegraphics[width=1\linewidth]{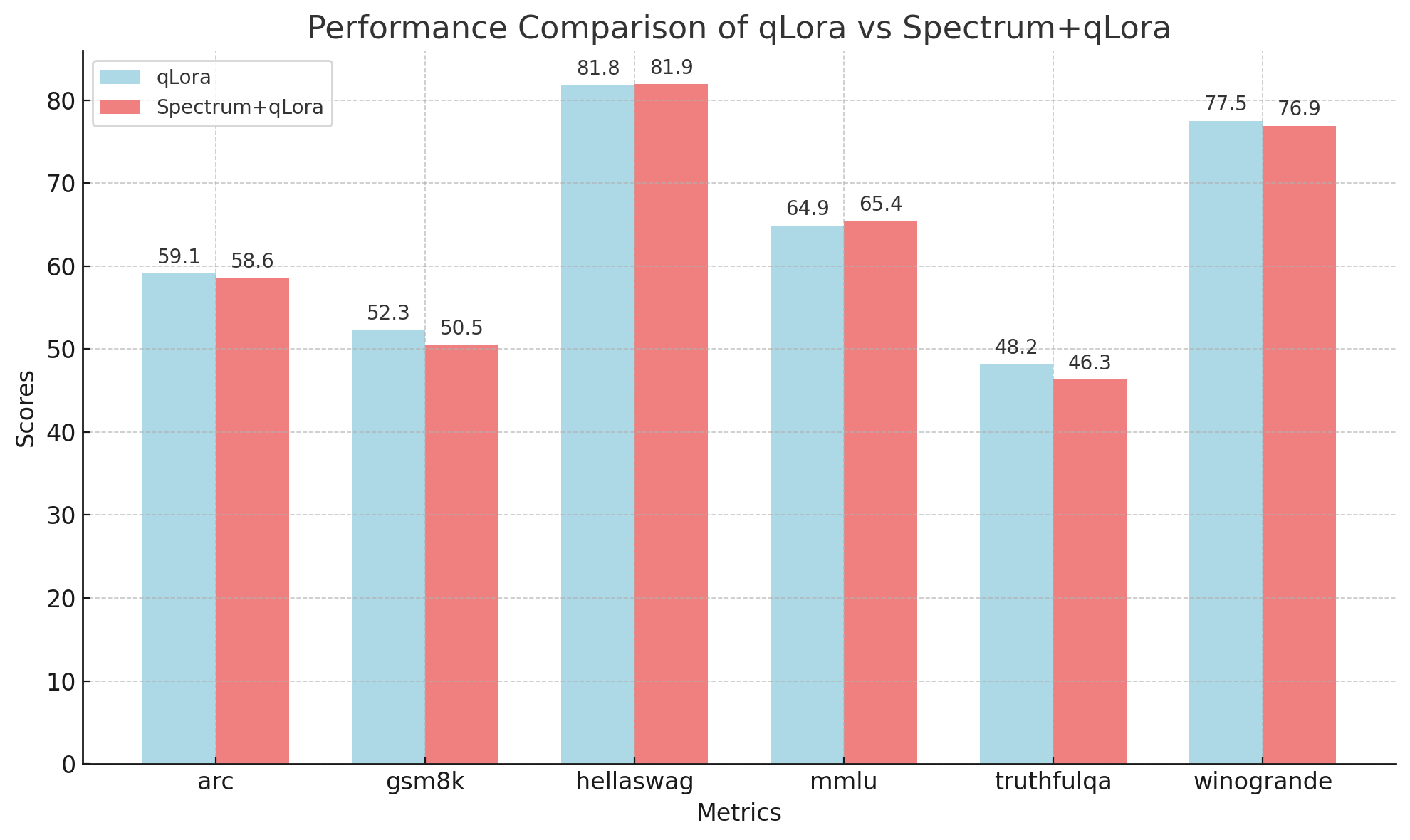}
    \label{fig:enter-label}
\end{figure}

\subsubsection{Memory Efficiency}

The efficiency of Spectrum is particularly evident when comparing it to QLoRA in distributed training settings using DeepSpeed ZeRO-3. Spectrum-50 and Spectrum-25 achieve 17.72\% and 23.05\% memory savings per GPU, respectively, compared to full finetuning. In contrast, QLoRA offers 14.73\% memory savings in distributed workloads.
\newline

QLoRA exhibits better memory efficiency than Spectrum when training on a single GPU. In the single GPU setting, most of the VRAM is used to load the model, reducing Spectrum's relevance. The efficiency gains for Spectrum come from only updating the gradients for the selected modules, which uses comparatively less memory than updating the entire model. This also explains why it is so much more efficient in distributed environments like DeepSpeed ZeRO-3 and Fully Sharded Data Parallel (FSDP) \cite{zhao2023fsdp}, as these allow us to train the model in a sharded manner across multiple GPUs, making the per GPU model memory footprint \textit{much} lower. Spectrum+QLoRA had the lowest VRAM requirement on both single and distributed workloads.

\subsubsection{Training Time}

In terms of training time, Spectrum demonstrates significant improvements over full finetuning and QLoRA (Table \ref{table:training_time}). Spectrum-50 and Spectrum-25 achieve 15.48\% and 36.78\% reductions in training time, respectively, compared to full finetuning. QLoRA also offers a 24.19\% reduction in training time. It should be noted that as the size of models grows, the time it takes to train using LoRA often increases. 

\section{Real World Use}

Spectrum has been used to fine-tune many Dolphin models. Fine-tuning large models like Qwen1.5-110B\footnote{\url{https://huggingface.co/Qwen/Qwen1.5-110B}} on an 8xH100 node (640 GB) encounters VRAM limitations. By using Spectrum-45, we were able to decrease the VRAM usage to 710 GB, fitting it within a single node while offloading some parameters to the CPU.

\begin{figure} [H]
    \centering
    \includegraphics[width=1\linewidth]{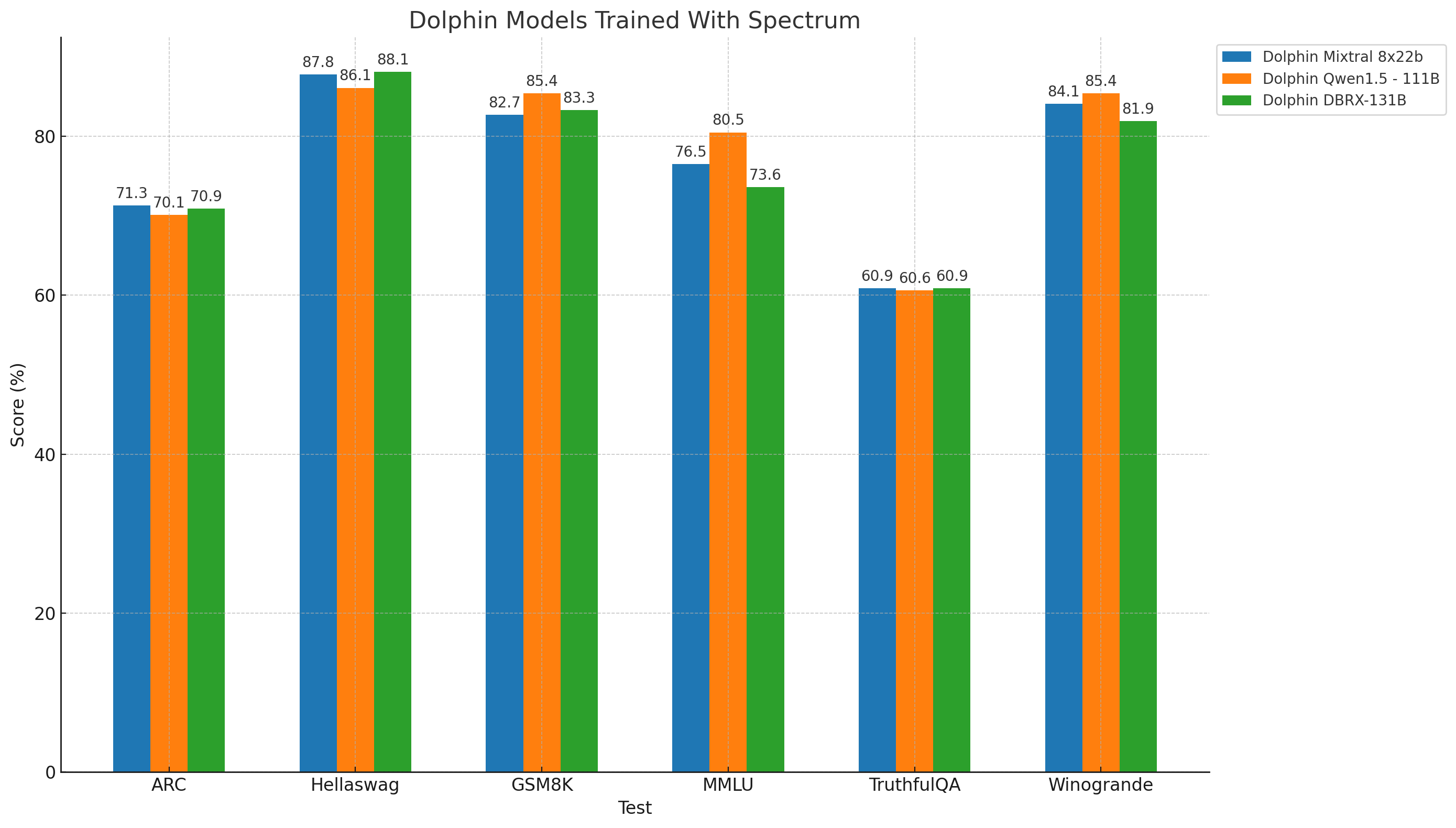}
\end{figure}

\section{Conclusion and Future Work}

We introduced Spectrum, a method for efficient training of LLMs. Spectrum selectively trains a subset of layers based on their signal-to-noise ratio (SNR), allowing it to focus compute on the most informative parameters. Spectrum uses significantly less memory in distributed workloads than prior methods while maintaining model quality.
\newline

Through evaluations on a set of language modeling benchmarks, we demonstrated Spectrum's performance compared to full finetuning and QLoRA. Spectrum opens up several promising directions for future work on efficient language model training. Spectrum's ability to identify the most informative layers can be further leveraged to enhance training efficiency. By employing layer-wise learning rate scheduling based on the signal-to-noise ratio (SNR), we can assign higher learning rates to the most informative layers. Additionally, dynamically rescanning the model between epochs and adjusting the targeting layers can further improve convergence speed. These techniques, currently under experimentation, aim to optimize the training process and accelerate model convergence.
\newline

The versatility of Spectrum extends beyond language modeling. Its ability to efficiently adapt pretrained models makes it well-suited for domain adaptation and transfer learning tasks. Furthermore, while our current work focuses on relatively small models, we have successfully applied Spectrum to train large language models with hundreds of billions of parameters. Exploring even larger models and datasets, as well as adapting Spectrum to other modalities offers exciting opportunities to broaden the impact of our work. We are excited about Spectrum's potential, and look forward to seeing how the community builds upon this foundation.
\newline

Spectrum will be made publicly available to facilitate adoption and future research\footnote{\url{https://github.com/cognitivecomputations/spectrum}}.

\subsection{Acknowledgements}

We would like to thank Crusoe Energy for providing the compute resources for our experiments. Special thanks to the teams working on Hugging Face Accelerate and OpenAccess AI Collective's Axolotl for their invaluable tools that made our research possible. We would also like to extend our heartfelt thanks to Maxime Labonne for being an early reviewer and for his honest and gracious feedback. Finally, we express our gratitude to the open-source community for their continuous contributions to democratizing access to machine learning tools.
\newline

\section*{Appendix}

We evaluated each model using \texttt{lm-evaluation-harness} commit \texttt{00b7a61} on multiple language modeling benchmarks popularized by the OpenLLM Leaderboard:

\begin{itemize}
    \item \textbf{Arc-Easy}: The AI2 Reasoning Challenge (ARC) Easy Set consists of 7,787 genuine grade-school level, multiple-choice science questions. The Easy Set includes questions that are straightforward for both humans and basic algorithms to answer correctly.

    \item \textbf{GSM8K}: (Grade School Math 8K) is a dataset of 8,500 high-quality, linguistically diverse grade school math word problems. These problems require multi-step reasoning and involve basic arithmetic operations.

    \item \textbf{HellaSwag}: Designed to evaluate commonsense natural language inference (NLI). It includes context completion tasks where models must choose the correct ending from multiple options, with adversarially generated incorrect endings.

    \item \textbf{MMLU}: The Massive Multitask Language Understanding (MMLU) benchmark consists of multiple-choice questions across 57 subjects, including STEM, humanities, social sciences, and more. It tests both world knowledge and problem-solving ability.

    \item \textbf{TruthfulQA}: A benchmark designed to measure the truthfulness of language models in generating answers to questions. It includes 817 questions across 38 categories, targeting common misconceptions and false beliefs.

    \item \textbf{Winogrande}: A dataset for evaluating commonsense reasoning, consisting of sentence pairs with a pronoun that needs to be resolved. The dataset is designed to be more challenging than the original Winograd Schema Challenge.
\end{itemize}

To further validate our results, we ran our Llama models on another suite of benchmarks popularized by Nous Research:
\newline
\begin{itemize}[noitemsep, topsep=0pt, parsep=0pt, partopsep=0pt] 
    \item \textbf{AGIEval}: A benchmark designed to evaluate the general intelligence of AI models. It includes a variety of tasks that test different aspects of reasoning, problem-solving, and knowledge application.
\newline

    \item \textbf{BigBench-Hard}: A subset of the BIG-Bench dataset, consisting of 23 particularly challenging tasks for current language models. These tasks require complex reasoning and understanding, often involving multi-step processes.
\newline

    \item \textbf{GPT4All}: A collection of benchmarks including BoolQ, PIQA, HellaSwag, WinoGrande, ARC-easy, ARC-c, and OBQA. (Note, there is some overlap here between these benchmarks and the OpenLLM Leaderboard tasks performed above.)
\end{itemize}

\vspace{1em} 

We used the chatml prompt template for each finetune, thus adding the \texttt{\textless{}im\_start\textgreater{}} and \texttt{\textless{}im\_end\textgreater{}} tokens to each model's vocabulary. This affects memory usage. By adding the same tokens to every finetune, we believe the comparisons in memory efficiency to be representative of what they would be without adding any additional tokens.
\newline

Additional Dolphin Models trained using Spectrum are referenced here. \footnote{\url{https://huggingface.co/cognitivecomputations/dolphin-2.9.2-mixtral-8x22b}}\footnote{\url{https://huggingface.co/cognitivecomputations/dolphin-2.9.1-qwen-110b}}\footnote{\url{https://huggingface.co/cognitivecomputations/dolphin-2.9.1-dbrx}}\footnote{\url{https://huggingface.co/cognitivecomputations/dolphin-2.9.1-llama-3-70b}}
\bibliographystyle{acm}
\bibliography{references}






\end{document}